\documentclass[9pt, nonacm, sigconf]{acmart}

\usepackage{pdfpages}
\usepackage{listings}
\usepackage{url}
\usepackage{multirow}
\usepackage{subfigure}
\usepackage{epsfig}
\usepackage{graphicx}
\usepackage{amsmath}
\usepackage{color}
\usepackage{setspace}
\usepackage{hhline}   
\usepackage{algorithm}
\usepackage{algpseudocode}
\usepackage{flushend}
\usepackage{url}
\usepackage[normalem]{ulem}
\usepackage{graphics}
\usepackage{tikz}
\usepackage{array}
\usepackage{animate}
\usepackage{titlesec}
\usepackage{balance}
\usepackage{enumitem}
\usepackage{booktabs}
\usepackage{hyperref}
\hypersetup{
  hidelinks,
  colorlinks,
  linkcolor={red!80!black},
  citecolor={blue!80!black},
  urlcolor={blue!80!black}
}

\titlespacing\section{2pt}{2pt plus 2pt minus 2pt}{2pt plus 2pt minus 2pt}
\titlespacing\subsection{2pt}{2pt plus 2pt minus 2pt}{2pt plus 2pt minus 2pt}
\titlespacing\subsubsection{2pt}{2pt plus 2pt minus 2pt}{2pt plus 2pt minus 2pt}
\titlespacing\paragraph{10pt}{2pt plus 2pt minus 2pt}{2pt plus 2pt minus 2pt}[]

\definecolor{codegreen}{rgb}{0,0.6,0}
\definecolor{codegray}{rgb}{0.5,0.5,0.5}
\definecolor{codepurple}{rgb}{0.58,0,0.82}
\definecolor{backcolour}{rgb}{0.95,0.95,0.92}
\definecolor{Burgundy1}{RGB}{159,29,53}
\lstdefinestyle{mystyle}{
    commentstyle=\color{codegreen}\itshape,
    keywordstyle=\color{black},
    numberstyle=\tiny\color{codegray},
    stringstyle=\color{Burgundy1},
    basicstyle=\scriptsize,
    breakatwhitespace=false,         
    breaklines=true,                 
    captionpos=b,                    
    keepspaces=true,                 
    backgroundcolor=\color{backcolour},   
    numbersep=2pt,                  
    showspaces=false,                
    showstringspaces=false,
    showtabs=false,                  
    tabsize=2
}

\begin{document}

\title{EvoDRC: A Self-Evolving Agentic Framework for \\Automated DRC Violation Repair}

\author{Bing-Yue Wu$^1$, Chia-Tung Ho$^2$, Haoyu Yang$^2$, Brucek Khailany$^2$, and Vidya A. Chhabria$^1$ \\ $^1$Arizona State University; $^2$NVIDIA Corporation}
\renewcommand{\shortauthors}{B.-Y. Wu, C.-T. Ho, H. Yang, B. Khailany, and V. A. Chhabria}

\renewcommand{\shorttitle}{}

\begin{abstract}
\noindent
Design rule check (DRC) closure remains a major bottleneck in advanced-node physical design. Although detailed routers are rule-aware, residual design rule violations (DRVs) often require manual engineering change order iterations. Automating this process is challenging because repairs must account for complex geometric interactions, preserve circuit connectivity, and avoid introducing new violations.
We present EvoDRC, a skill-evolution framework for agentic block-level DRC repair. EvoDRC initializes layer-specific repair skills using knowledge distilled from an unrelated reference design and continuously evolves these skills using traceable repair experience collected from the target design. EvoDRC decomposes the layout into bounded repair regions and assigns an LLM repair agent to each region. Local DRC analysis, connectivity-checking, and impact-preview tools provide feedback on proposed modifications. Repair operations and their resulting DRV changes are stored in a knowledge database and used to evolve the repair skills. Experiments on seven block-level designs from the DAC26 DRC Benchmark show that EvoDRC achieves a 73.5\% overall reduction compared to the reported baseline.

\end{abstract}	
\maketitle

\section{Introduction} 
\label{sec:introduction}
\noindent
Design rule check (DRC) closure remains a critical bottleneck in advanced-node physical design. Foundry design rules impose increasingly complex geometric constraints involving metal width, spacing, enclosure, alignment, and interactions across multiple routing layers. Although modern detailed routers are rule-aware, they do not always produce DRC-clean layouts. Residual design rule violations (DRVs) must therefore be resolved through engineering change order (ECO) iterations, which often require substantial manual effort from layout engineers. Engineers must inspect each violation in its surrounding layout context, identify the relevant design rule, determine its root cause, and apply a repair without disrupting circuit connectivity or introducing additional violations. This process is inherently iterative because a local geometric edit can produce ripple effects elsewhere in the layout. Consequently, post-route DRC repair can consume significant engineering effort. Industry reports indicate that iterative DRC closure can require multiple
ECO and verification cycles over several days or weeks%
~\cite{semiengineering}.

Large language models (LLMs) have recently demonstrated strong capabilities as agents that interact with external tools, reason over intermediate results, and refine their actions in a closed loop~\cite{yao2023react}. Recent studies in physical design use agentic systems to generate scripts and interact with EDA tools for automating placement, routing, and timing optimization~\cite{openroad_assistant,openroad_agent,orfs_agent,wu2024chateda,generative_methods_eda,CAPO}. These results suggest that LLM agents can automate key physical design stages. However, agentic DRC repair remains largely unexplored. Existing studies apply LLMs to design-rule interpretation and DRC checker code generation~\cite{drc_coder}, while recent efforts demonstrate agentic DRC repair on small-scale tasks, such as standard-cell-level layouts~\cite{2026dacdrc,liu2026bridgingmilecircuitdesign}. These methods are difficult to scale to block-level complex designs with diverse design rules spanning multiple routing layers. Moreover, an LLM agent's repair capability strongly depends on a well-developed, task-specific skill file that encodes factual repair knowledge, but such a skill file is rarely available for a new target design.

\begin{figure}[t]
    \centering
    \includegraphics[width= 0.89\linewidth]{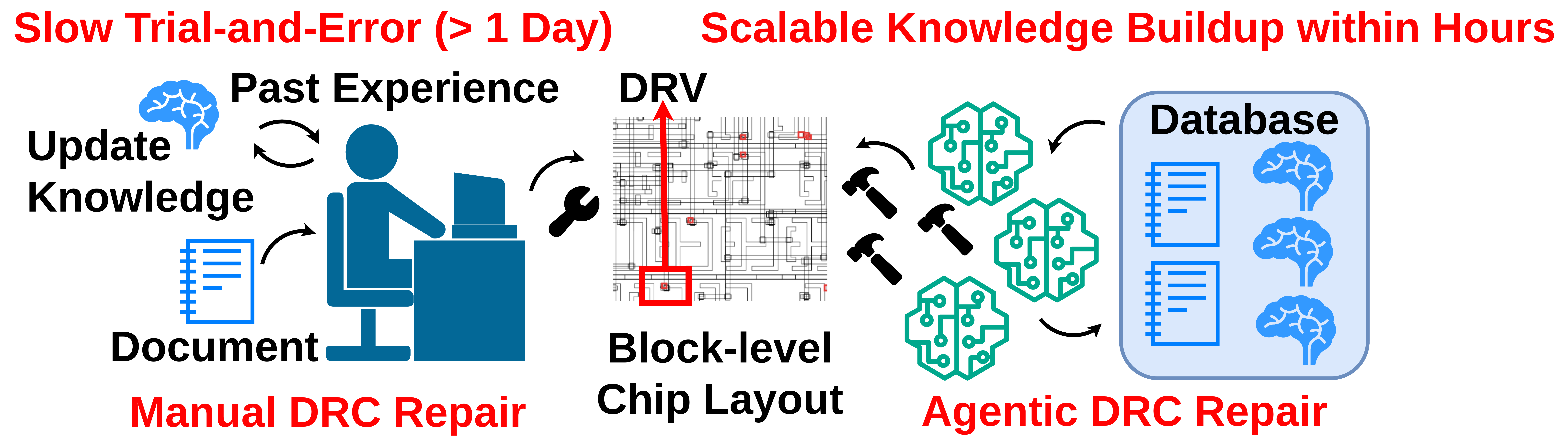}
    \vspace{-13pt}
    \caption{Human vs agent in DRC repair.}
    \vspace{-16pt}
    \label{fig:intro}
\end{figure}

In practice, a DRC engineer often performs similar repair tasks across multiple designs, as shown in Fig.~\ref{fig:intro}. When a familiar violation pattern appears, the engineer can reuse repair strategies learned while transforming previous DRC-violating designs into DRC-clean designs. However, a new layout may expose previously unseen rule interactions or geometric contexts, requiring the engineer to refine existing strategies or learn new ones. We formulate this process as a \emph{skill-evolution} task: repair experience from a previous design provides an initial set of transferable skills, while verified repair outcomes from the target design are used to continuously evolve those skills. In this work, we present \textbf{EvoDRC}, a skill-evolution framework targeting block-level DRC repair.
EvoDRC initializes its repair skills by distilling knowledge from transforming an unrelated DRC-violating reference design into a DRC-clean design. It subsequently refines these skills using traceable repair experience collected during repairs to the target design. To improve scalability, EvoDRC decomposes large layouts into bounded \textit{repair crops} and dispatches one repair agent per crop. Each agent explores repair strategies using preview tools that provide feedback on local DRC changes, circuit connectivity, and the impact of a proposed edit. EvoDRC stores approved repair operations together with their resulting changes in a knowledge database (Knowledge DB). It then evolves layer-specific skill files over successive repair iterations by distilling both successful and unsuccessful repair records. This process combines cross-design skill transfer with test-time skill evolution while preserving a traceable connection between each learned skill and its repair outcomes.  The contributions are:

\begin{itemize}[nosep, leftmargin=*]
    \item To the best of our knowledge, EvoDRC is the first agentic DRV repair framework that automates traditionally manual post-route DRC ECOs, allowing layout engineers to focus on other tasks.
    
        \item EvoDRC formulates DRC repair as a skill-evolution task. By combining skill transfer from a reference design with skill evolution on the target design, EvoDRC continuously improves the repair knowledge provided to its agents.

    \item EvoDRC introduces a layout-decomposition mechanism that improves the scalability of agentic DRC repair by dividing block-level layouts into bounded repair crops.


    \item Experimental results on block-level test cases from the DAC26-DRC-Benchmark~\cite{DAC26_DRC_Benchmarks} show that EvoDRC reduces DRV by 73.5\%.
\end{itemize}

\noindent
EvoDRC is available at~\cite{EvoDRC}.

\section{Related work}
\label{sec:background}


\subsection{Placer- and Router-Driven DRC Reduction}
\label{subsec:pnr-drc}
\noindent
Most existing efforts focus on reducing DRVs during the placement and routing stages of the physical design flow. At the placement stage, predictive methods estimate DRV hotspots before routing and reshape the layout through methods such as synthesized routing blockages, whitespace redistribution, or routing resource adjustment~\cite{10.1145/3676536.3676828,10.1145/3626184.3633324, 10.1145/3240765.3240843}. However, these techniques still leave residual DRVs that must be repaired manually. At the routing stage, rule-aware detailed routers integrate DRC into itself through pin-access analysis, track assignment, and adaptive rip-up~\cite{9120211,10.1145/3526241.3530361}. In practice, the router checks a simplified rule model instead of the sign-off deck for scalability, so layouts still have residual DRVs. After routing, ECO-stage optimizers repair residual DRVs by jointly refining cell placement, routing wires, and cell selection within a local region under satisfiability modulo theories (SMT).~\cite{9643792,10528730}. However, each rule must be hand-translated into constraints, so these optimizers do not extend to a full sign-off deck.

\subsection{LLMs for DRC-related tasks}
\label{subsec:llm-drc}
\noindent
Recent works have explored the application of LLMs to DRC-related tasks. Existing studies use LLM agents with multimodal design-rule interpretation, code execution, and debugging feedback to translate design-rule descriptions into executable DRC checker code~\cite{drc_coder}. The results demonstrate that LLMs can read design-rule descriptions, relate them to layout geometry, and produce correct checking logic. However, generating a checker addresses violation detection rather than violation repair. In DRC checker code generation, the agent neither modifies the layout nor encounters the ripple effects of geometric edits. Agentic DRC repair, therefore, requires a different interface, in which the agent proposes layout operations and an external framework verifies their consequences.

\subsection{DRC Repair Benchmarks}
\label{subsec:drc-benchmarks}
\noindent
Open-source benchmarks for DRC repair have only recently appeared~\cite {2026dacdrc,DAC26_DRC_Benchmarks,liu2026bridgingmilecircuitdesign,PostEDA-Bench}. One representative benchmark, the DAC26-DRC-Benchmark~\cite{2026dacdrc, DAC26_DRC_Benchmarks}, releases test cases at multiple scales, including polygon-level, standard-cell-level, and block-level cases, to evaluate both the capability and scalability of LLM workflows on DRC repair problems of varying scale of problem. Implemented with the ASAP7 PDK~\cite{asap7pdk}, this benchmark provides each test case with the layout GDSII, a layout script, a DRC report from KLayout, the open-source DRC tool~\cite{klayout_website}, and a net connectivity checker. Its experiments provide an agentic DRC repair workflow with the layout GDSII, layout script, design rule deck, design rule manual (DRM) information, and KLayout tool. The results show that many available LLMs exhibit strong reasoning capability on standard-cell-level DRC repair tasks. However, repair quality degrades drastically when the same workflow is applied to block-level layouts. This indicates the need for a scalable approach to deploying agentic DRC repair on block-level cases. Another benchmark, PostEDA-Bench~\cite{liu2026bridgingmilecircuitdesign,PostEDA-Bench}, provides checkable tasks for residual sign-off DRC repair and reveals a substantial capability gap between synthetic rule-level exercises and practical geometric, multi-step DRC reasoning. Both benchmarks provide evaluation harnesses, making agentic DRC repair directly measurable and reproducible. In this work, we evaluate EvoDRC on the block-level test cases from the DAC26-DRC-Benchmark rather than PostEDA-Bench, since PostEDA-Bench is limited to a single violation and its DRC-fixing tasks are restricted to small designs with at most 14 DRVs, whereas the DAC26-DRC-Benchmark blocks contain up to 765 DRVs.

\section{The EvoDRC Framework}
\label{sec:framework}

\begin{figure*}[t]
    \centering
    \vspace{-24pt}
    \includegraphics[width= 1\linewidth]{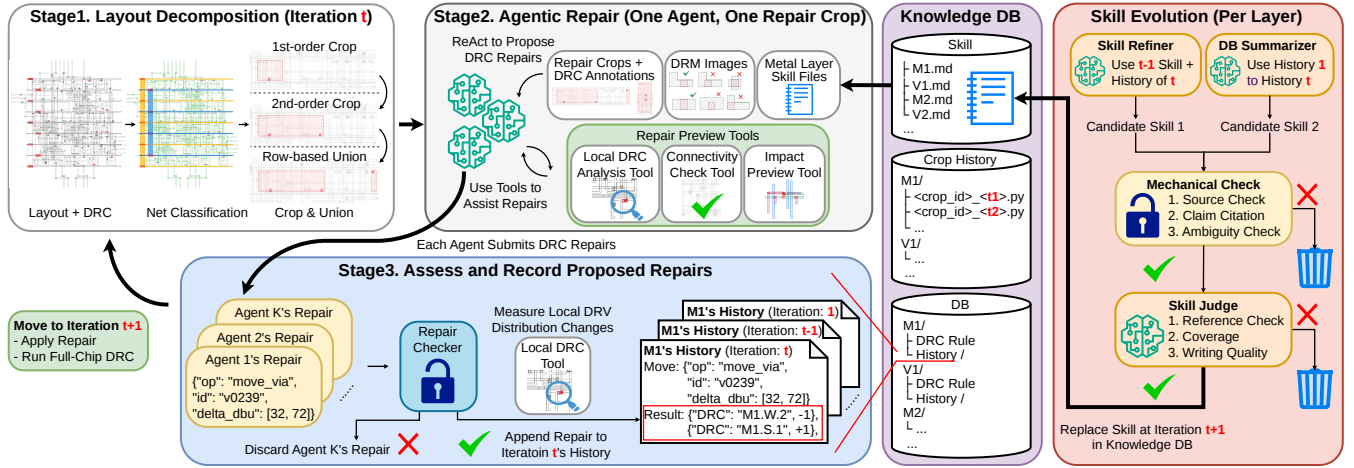}
    \vspace{-22pt}
    \caption{Overview of EvoDRC. The framework consists of layout decomposition, agentic DRC repair, and skill evolution. Stage~1 decomposes the layout into repair crops based on the DRV distribution (marked in red). Stage~2 dispatches agents to repair each crop using preview tools, DRM images, and layer-based skill files. Stage~3 records validated operation-result pairs in the Knowledge DB to evolve skills for the next iteration.}
    \vspace{-10pt}
    \label{fig:evodrc}
\end{figure*}

\noindent
Fig.~\ref{fig:evodrc} illustrates the overview of the EvoDRC framework. First, in the layout decomposition stage (left top corner of Fig.~\ref{fig:evodrc}), given a layout and its DRC report, EvoDRC classifies different routing types and then crops the layout based on the DRV distribution for each routing type to form \textit{repair crops}. Second, in the agentic repair stage, DRC repair agents are dispatched to work on each repair crop. Additional tools are provided to assist the repair process by informing the agents of the consequences of intermediate repair trials, including intermediate DRV distribution changes, net connectivity preservation, and the impact range of each action. The DRC repair agents are provided with the assigned repair crop and the DRV distribution of that crop, DRM images as visual cues for the design rules, and layer-based skill files that document the design rule deck used to run DRC analysis and the DRC repair knowledge used to guide layer-wise DRC repair. The agents adopt the ReAct paradigm~\cite{yao2023react} to reason and act on DRC repair.

In the third stage, after the agents provide the suggested repair operations, EvoDRC examines whether the operations break circuit logic and decides which operation to take when conflicting operations from different agents exist. After examination, the accepted repair operations are applied to each corresponding repair crop. EvoDRC then runs DRC on each crop and records the operation-result pairs into the knowledge database (Knowledge DB). In the Knowledge DB, two agents, i.e., the Skill Refiner and the DB Summarizer, are dispatched to generate candidate skill files for each layer that may be used in the next iteration of DRC repair through two different approaches. The generated skill files are then passed through a script-based checker to ensure quality. The Skill Judge agent then takes the candidate skill files and decides whether the current skill file in the Knowledge DB should be replaced. EvoDRC-driven DRC repair then proceeds to the next iteration with the updated skill file. EvoDRC is designed to refine and evolve the skill file that guides DRC repair over iterations, helping DRC repair agents achieve higher repair quality.

\subsection{Layout Decomposition}
\label{subsec:layout-decomposition}
\noindent
As shown in Stage~$1$ in Fig.~\ref{fig:evodrc}, EvoDRC first parses the design and classifies different routing types. For each routing type, EvoDRC builds layout crops based on the DRV distribution from the DRC report. Regular metal-wire polygons and vias that touch or overlap the DRV region are used to enlarge the cropped layout by computing the minimum and maximum vertex coordinates in the x and y directions to form a first-order crop. To prevent a long stripe from expanding a crop to an unnecessarily large crop, EvoDRC bounds the first-order crop to a size of \(k\,\mu\mathrm{m}\times k\,\mu\mathrm{m}\). First-order crops that overlap in x and y coordinates and metal layers are merged into second-order crops. EvoDRC then unions second-order crops that fit within the same standard-cell-height row to form final \textit{repair crops}, while leaving each multi-row crop as an independent repair crop. The resulting repair crops are then passed individually to the repair agents in parallel.

\subsection{Agentic Repair}
\label{subsec:agentic-repair}
\noindent
As Stage~$2$ in Fig.~\ref{fig:evodrc} illustrates, for each repair crop, the repair agent receives (i) design rule manual (DRM) images, (ii) the repair crop represented by a KLayout Python script, (iii) its DRV distribution, i.e., DRV type and location, and (iv) layer-wise skill files for the layers spanned by the repair crop. Each layer-wise skill file combines the corresponding KLayout DRC deck describing the design rules with repair knowledge distilled from previous repair iterations. 

\begin{table}[t]
    \centering
    \vspace{-3pt}
    \caption{Object editability classes and available operations.}
    \vspace{-12pt}
    \label{table:class_operation}
    \fontsize{7pt}{8pt}\selectfont
    {\setlength{\tabcolsep}{2pt}
    \renewcommand{\arraystretch}{0.9}
    \begin{tabular}{|c|m{3cm}|m{1.7cm}|m{2.6cm}|}
        \hline
        \textbf{Class} &
        \textbf{Object type} &
        \textbf{Operation} &
        \textbf{Applicable scope} \\ \hline

        \multirow{5}{*}{\textbf{A}}
        & \multirow{5}{2cm}{Via structure}
        & move\_via & Move a via \\ \cline{3-4}
        & & delete\_via & Delete a via \\ \cline{3-4}
        & & add\_via & Add a via \\ \cline{3-4}
        & & resize\_via\_shape & Resize a metal polygon within a via structure \\ \cline{3-4}
        & & move\_via\_shape & Move a metal polygon within a via structure \\ \hline

        \multirow{5}{*}{\textbf{B}}
        & \multirow{5}{1.7cm}{Polygon fully within the region}
        & resize & Resize a polygon \\ \cline{3-4}
        & & move & Move a polygon \\ \cline{3-4}
        & & delete & Delete a polygon \\ \cline{3-4}
        & & add\_vertex & Modify the shape of a polygon \\ \cline{3-4}
        & & add\_polygon & Add a polygon \\ \hline

        \textbf{C}
        & Polygon cropped by the \(k\,\mu\mathrm{m}\) window
        & resize\_end
        & Shorten or elongate the open end \\ \hline

        \textbf{D}
        & Standard cell and polygons of other routing types
        &
        & No repair operation allowed \\ \hline
    \end{tabular}}
\end{table}

\begin{table}[t]
    \centering
    \vspace{-11pt}
    \caption{Preview tools available during repair.}
    \vspace{-12pt}
    \label{table:preview_tool}
    \fontsize{7pt}{10pt}\selectfont
    {\setlength{\tabcolsep}{3pt}
    \renewcommand{\arraystretch}{0.85}
    \begin{tabular}{|m{0.20\columnwidth}|m{0.75\columnwidth}|}
        \hline
        \textbf{Tool} & \textbf{Returned information} \\ \hline
        \textbf{Local DRC analysis} & Run DRC on the selected repair region in a sandbox for fast region-local feedback. Reports whether the target DRV is cleared, remaining targets, and new DRVs with bounding boxes \\ \hline
        \textbf{Connectivity check} & Result of the full-design connectivity check and the seeds of broken nets \\ \hline
        \textbf{Impact preview} & Metal/via objects touching or overlapping an edited object, including each neighbor's layer, kind, and editability \\ \hline
    \end{tabular}}
    \vspace{-15pt}
\end{table}

EvoDRC assigns each metal polygon or via structure in the repair crop to one of the four classes in Table~\ref{table:class_operation}: (i) Vias in the repair crop can be moved, deleted, or added only as complete objects. No partial edits may be applied to a single via. Any modification to a via structure must be applied to all vias in the whole layout. (ii) Metal stripes fully within the repair crop are allowed to have any operation applied to them, such as shape modification, moving, deleting, or creating new metal stripes. (iii) Cropped metal stripes in the repair crop are restricted to elongating or shortening their open ends within the repair crop to prevent unexpected consequences in unseen regions of the layout. (iv) Standard cells internal polygons are not allowed to be modified to prevent degradation of the circuit logic, as are metal stripes and vias that belong to other routing types. These classes distinguish editable objects from frozen context and prevent a crop boundary from implicitly granting control over an unseen portion of the design. Table~\ref{table:class_operation} defines the available operations for each class of object.

To assist the repair agent in its ReAct-based agentic repair, three preview tools in Table~\ref{table:preview_tool} are provided to each repair agent: (i) The local DRC analysis tool can be used to run DRC analysis on the repair crop and show the DRV distribution difference during repair operation exploration, which helps the repair agent reason about different repair strategies. (ii) The connectivity check tool can verify whether the proposed repair operation breaks net connectivity, which affects circuit logic, so such operations are forbidden\footnote{This check is not an LVS run. It applies the candidate operations to a copy of the layout and reruns the DAC26-DRC-Benchmark connectivity checker, a simple depth-first search over the merged metal--via stack that verifies each standard-cell pin still reaches its corresponding standard-cell or IO pins.}. (iii) The impact preview tool is used to provide the possibly affected objects when the agent tries to apply operations to objects in the repair crop. Each repair agent repairs the assigned repair crop using the provided input and tools, and returns a list of repair operations.

\subsection{Assess and Record Proposed Repairs}
\label{subsec:record-repair}
\noindent
EvoDRC accepts a repair agent's final repair operation list only when it passes the full-design connectivity check. The framework applies this list of repair operations to the repair crop and runs local DRC on the repair crop to record the local DRV distribution changes. When conflicting operations are proposed by different repair agents, e.g., when two repair agents propose different approaches to modifying the structure of the same generated via, the same edit on the same object may have different impacts in different contexts because each repair agent only reasons about and proposes repair operations to resolve local DRVs. Therefore, for repair crops whose agents propose structure edits on the same generated via, EvoDRC applies the via edits individually and tests which via edit performs best, i.e., yields the most net DRV count reduction or the smallest DRC increase.

Finally, EvoDRC groups the repair operation--DRV distribution change records by metal layer and appends them to the Knowledge DB, as shown in Stage~3 in Fig.~\ref{fig:evodrc}. Each record includes the repair operations on each layer and their corresponding DRV distribution changes on the same metal layer, preserving trackable results rather than blindly trusting an agent's output. The geometry of repair crops is also stored in the Knowledge DB for future replay purposes.

\subsection{Knowledge DB}
\label{subsec:knowledge-db}
\noindent
The key component of EvoDRC is the Knowledge DB, which not only stores the skill files for all layers used in the agentic DRC repair process but also keeps the proposed repair operations and DRV distribution changes within the repair crop, as well as the geometry of the repair crops. The Knowledge DB stores these records by metal layer and iteration timestamp.

\subsubsection{Knowledge DB Organization}
The purple box in Fig.~\ref{fig:evodrc} summarizes the database organization. The Knowledge DB stores three types of data: (i) the \texttt{DB} directory contains storage directories for each metal layer, such as M1, M2, and V1. Each directory stores the \texttt{Design Rule}, which is the corresponding KLayout DRC deck for that layer describing the design rule, and the \texttt{History} directory, which stores all repair operation--local DRV distribution change records. (ii) the \texttt{Skill} directory stores the skill file for each metal layer, while the \texttt{Crop History} directory preserves repair crop geometry by layer, crop, and iteration annotation, serving as a backup to replay the geometry of repair crops from previous iterations.

\subsubsection{Trackable Records}
\label{subsec:trackable-records}
EvoDRC is intentionally designed to have each repair agent return only its final proposed repair operations in order to keep all records trackable. Intermediate trials are often untraceable because they may start from temporary geometries produced by earlier speculative edits during the agents' DRC repair exploration. Simply applying such operations to the original geometry can cause them to be accumulated as noisy data and progressively bias the Knowledge DB. As a result, EvoDRC discards all intermediate trial records and only allows each repair agent to return its final verdict, which is then stored in the Knowledge DB.

\subsection{Skill Evolution}
\label{subsec:skill-generation}
\noindent
After the operation--DRV distribution change records of the current DRC repair iteration are stored, EvoDRC launches two skill-evolution agents independently for each metal layer with new records, as shown on the right of Fig.~\ref{fig:evodrc}. (i) The Skill Refiner reads the \texttt{Design Rule} and the skill file for the corresponding layer from the Knowledge DB and combines them with the records collected in the current iteration. (ii) The DB Summarizer reads the complete history of that layer across all iterations, along with the \texttt{Design Rule} from the Knowledge DB, to construct a skill file for that layer.
The Skill Refiner incrementally refines the existing skill files using the repair operation and DRV distribution change records collected in the current iteration, whereas the DB Summarizer reconstructs a new skill file from that layer’s complete repair history. Both knowledge skill candidates are evaluated against the same metrics described in Sections~\ref{subsec:mechanical-check} and~\ref{subsec:skill-judge}. Example skill files are at~\cite{EvoDRC}. 

\subsubsection{Mechanical Check}
\label{subsec:mechanical-check}
After the Skill Refiner and the DB Summarizer each generate a skill file, EvoDRC applies three script-based checks to each skill file candidate:
\begin{enumerate}[nosep, leftmargin=*]
    \item Source check: Every reference must exist in the layer's \texttt{History} directory in the Knowledge DB.
    \item Claim citation: Every paragraph in the skill candidate having assertive words such as ``do not,'' ``never,'' or ``always,'' must at least cite one existing reference in the layer's \texttt{History} directory in the Knowledge DB.
    \item Ambiguity check: Must not use uncertain words such as `` likely,'' ``maybe,'' ``might,'' or ``untested.'' 
\end{enumerate}
Candidates that fail any check are discarded. These checks enforce minimum traceability rather than factual correctness.

\subsubsection{Skill Judge}
\label{subsec:skill-judge}
When both skill candidates pass the mechanical checks, the Skill Judge agent compares them using three metrics:
\begin{enumerate}[nosep, leftmargin=*]
    \item Reference check: Each statement must be backed by at least one repair record.
    \item Coverage: Prefers the candidate that accounts for more of the layer's repair records, weighting the newest most.
    \item Writing quality: Favors the candidate with less redundancy, clearer wording, and fewer contradictions.
\end{enumerate}
The candidate receiving the majority of the three votes is selected. If only one candidate passes the mechanical checks, the Skill Judge decides whether it should replace the current skill file. Otherwise, the existing skill is retained. The selected skill is stored in the \texttt{Skill} directory and used in iteration $t+1$ .EvoDRC then
performs full-layer DRC analysis and advances to iteration $t+1$, where the updated
skill is provided to the repair agents.

\section{Experimental Setup}
\label{sec:experiments}

\subsection{Benchmark and Test Cases}
\label{subsec:setup-testcases}
\noindent
To evaluate the scalability and effectiveness of the EvoDRC framework, we conduct experiments on the seven block-level test cases from the DAC26-DRC-Benchmark~\cite{DAC26_DRC_Benchmarks}. All cases are implemented with the ASAP7 PDK~\cite{asap7pdk}. Each test case provides the layout in the form of a GDSII file and a KLayout Python script, a KLayout DRC report, and net connectivity checker file of the design. We follow the DAC26-DRC-Benchmark and use KLayout~0.30.1~\cite{klayout_website} with the provided ASAP7 KLayout rule deck. The seven blocks span $68$ to $765$ initial DRVs, covering metal width, spacing, area, enclosure, and grid-alignment constraints. In addition to using these seven blocks as test cases, we implement an external design, ``carry\_lookahead\_adder'' (CLA), from OpenCores~\cite{OpenCores} using OpenROAD-flow-scripts~\cite{openroad-flow-scripts} and run DRC. Table~\ref{table:case_geo} shows the dimension and scale differences among all eight cases, and Table~\ref{table:case_drv} shows the DRC comparison across the cases. To the best of our knowledge, no prior work has attempted automatic, ECO-style DRC repair at this scale, and the DAC26-DRC-Benchmark, whose block-level design contains upto 765 DRVs, is the largest DRC repair benchmark available.

\begin{table}[t]
    \centering
    \vspace{-15pt}
    \caption{Benchmark and CLA design characteristics.}
    \label{table:case_geo}
    \vspace{-12pt}
    \fontsize{7pt}{8pt}\selectfont
    {\setlength{\tabcolsep}{3pt}
    \begin{tabular}{|l|r|r|r|r|c|c|}
        \hline
\textbf{Design} & \textbf{\#Insts.} & \textbf{Nets} &
\shortstack{\textbf{Die area}\\\textbf{($\mu\mathrm{m}^2$)}} &
\shortstack{\textbf{Core area}\\\textbf{($\mu\mathrm{m}^2$)}} &
\shortstack{\textbf{Die $W\times H$}\\\textbf{($\mu\mathrm{m}$)}} &
\shortstack{\textbf{Core $W\times H$}\\\textbf{($\mu\mathrm{m}$)}} \\
\hline
        \textbf{Block1} & 143 & 120 & 16.06 & 9.30 & \(4.008\times4.008\) & \(3.132\times2.970\) \\ \hline
        \textbf{Block2} & 62  & 43  & 8.99  & 4.08 & \(2.998\times2.998\) & \(2.160\times1.890\) \\ \hline
        \textbf{Block3} & 76  & 62  & 10.07 & 5.02 & \(3.174\times3.174\) & \(2.322\times2.160\) \\ \hline
        \textbf{Block4} & 107 & 81  & 13.45 & 7.58 & \(3.668\times3.668\) & \(2.808\times2.700\) \\ \hline
        \textbf{Block5} & 47  & 28  & 7.27  & 2.97 & \(2.696\times2.696\) & \(1.836\times1.620\) \\ \hline
        \textbf{Block6} & 155 & 101 & 17.94 & 11.02 & \(4.236\times4.236\) & \(3.402\times3.240\) \\ \hline
        \textbf{Block7} & 574 & 410 & 57.91 & 43.74 & \(7.610\times7.610\) & \(6.750\times6.480\) \\ \hline
        \textbf{CLA}    & 52  & 229 & 4.818 & 2.712 & \(2.195\times2.195\) & \(1.674\times1.620\) \\ \hline
    \end{tabular}}
    \vspace{-10pt}
\end{table}

\begin{table}[t]
    \centering
    \vspace{-1pt}
    \caption{Initial DRV distribution by rule type.}
    \vspace{-12pt}
    \label{table:case_drv}
    \fontsize{6pt}{9pt}\selectfont
    \renewcommand{\arraystretch}{0.85}
    {\setlength{\tabcolsep}{2.5pt}
    \begin{tabular}{|l|r|r|r|r|r|r|r|r|}
        \hline
        \multicolumn{1}{|c|}{\textbf{Rule}} & \textbf{Block1} & \textbf{Block2} & \textbf{Block3} & \textbf{Block4} & \textbf{Block5} & \textbf{Block6} & \textbf{Block7} & \textbf{CLA} \\ \hline
        \textbf{M4.AUX.X}    & 20 & 7  & 7  & 13 & 6  & 22 & 63  & 7 \\ \hline
        \textbf{V1.M1.EN.1}  & 11 & 2  & 6  & 4  & 5  & 10 & 26  & 7 \\ \hline
        \textbf{V3.M4.AUX.2} & -- & -- & -- & -- & -- & -- & --  & 7 \\ \hline
        \textbf{V0.M1.AUX.3} & 37 & 12 & 21 & 20 & 9  & 21 & 97  & 6 \\ \hline
        \textbf{M1.S.X}      & 14 & 2  & 6  & 8  & 8  & 17 & 94  & 5 \\ \hline
        \textbf{M5.AUX.X}    & 8  & 4  & 4  & 6  & 4  & 8  & 12  & 4 \\ \hline
        \textbf{M4.S.X}      & 4  & 1  & -- & 3  & -- & -- & 1   & 1 \\ \hline
        \textbf{V2.M3.AUX.2} & 72 & 24 & 27 & 51 & 21 & 78 & 225 & --\\ \hline
        \textbf{V4.M5.AUX.2} & 48 & 16 & 18 & 34 & 14 & 52 & 150 & --\\ \hline
        \textbf{V5.M6.AUX.2} & 24 & -- & -- & 6  & -- & 32 & 72  & --\\ \hline
        \textbf{M3.S.X}      & 2  & -- & -- & 1  & 1  & 3  & 15  & --\\ \hline
        \textbf{M6.AUX.X}    & 3  & -- & -- & 1  & -- & 4  & 6   & --\\ \hline
        \textbf{M2.S.X}      & 1  & -- & -- & -- & -- & -- & 4   & --\\ \hline
        \textbf{Rule types}  & 12 & 8  & 7  & 11 & 8  & 10 & 12  & 7 \\ \hline
        \textbf{Total DRVs}  & 244& 68 & 89 & 147& 68 & 247& 765 & 37\\ \hline
    \end{tabular}}\\[2pt]
    {\raggedright We group DRCs of the same type under a shared label, e.g., M2.S.X is spacing rules on M2, and M5.AUX.X is auxiliary rules on M5. Hyphen implies no violatoin of that type\par}

    \vspace{-10pt}
\end{table}

As shown in Table~\ref{table:case_geo} and Table~\ref{table:case_drv}, CLA is significantly smaller than the seven blocks in DAC26-DRC-Benchmark, and its initial DRV distribution differs substantially from theirs. We manually remove the DRVs in the CLA block and use the DB Summarizer agent described in Sec.~\ref{subsec:skill-generation} to generate the initial skill file, which serves as the input in the first iteration of DRC repair for all the seven test cases in the following experiments.

\subsection{Experiment Configuration}
\label{subsec:setup-agent}
\noindent
We use Claude Sonnet 4.6 from Anthropic as the agent~\cite{claude46_sonnet, anthropic_website} for all agents in the EvoDRC framework and set the reasoning effort to medium in the following DRC repair experiments. For the repair agent, it follows the ReAct paradigm and internally executes the four-role workflow: a planner, an adversarial plan reviewer, a DRC engineer, and a patch reviewer. As described in Sec.~\ref{subsec:agentic-repair}, each repair agent reads the assigned KLayout Python script-described repair crop, its DRV distribution, DRM images, and layer-wise skill files, and may use the three preview tools to assist DRC repair. There's no runtime limitation for agents to repair DRC, a repair task completes when the agent submits its final repair operation list. Example prompts for all agents can be found at~\cite{EvoDRC}.

We allow EvoDRC to run for at most five iterations and stop early when the block reaches zero violations. For the crop-size restriction described in Sec.~\ref{subsec:layout-decomposition}, we set it to \(2\,\mu\mathrm{m}\times 2\,\mu\mathrm{m}\). We quantify DRC repair quality using \% improvement in the net number of DRVs:
\begin{equation}
\mathrm{Improvement} =
\frac{\mathrm{DRV}_{\mathrm{initial}} - \mathrm{DRV}_{\mathrm{final}}}
{\mathrm{DRV}_{\mathrm{initial}}}
\times 100
\label{eq:drc_improvement}
\end{equation}
Here, $\mathrm{DRV}_{\mathrm{initial}}$ is the initial number of DRVs for each block, as shown in Table~\ref{table:case_drv}, and $\mathrm{DRV}_{\mathrm{final}}$ is the final number of DRVs after the experiment. As ablation, we compare three configurations each of which isolate one mechanism of EvoDRC:
\begin{itemize}[nosep, leftmargin=*]
\item \textbf{Ablation~1:} Remove the CLA-derived skill files in the first iteration, leaving only the KLayout DRC deck. This evaluates whether the initial skill files contribute to repair quality.
\item \textbf{Ablation~2:} EvoDRC using CLA-derived skill files for all iterations, with skill evolution disabled. This setup evaluates whether skill evolution contributes to repair quality.
\item \textbf{Ablation~3:} EvoDRC using CLA-derived skill files for the first iteration, with the Layout Decomposition stage disabled. This setup evaluates whether decomposing a large design into small repair crops contributes to repair quality.
\end{itemize}



\section{Results}
\label{sec:results}


\begin{table*}[t]
\centering
\vspace{-14pt}
\caption{DRV count after each repair iteration for EvoDRC and the three ablation studies.}
\vspace{-12pt}
\fontsize{6pt}{10pt}\selectfont
\renewcommand{\arraystretch}{0.85}
\label{table:results}
\setlength{\tabcolsep}{3pt}
\resizebox{\textwidth}{!}{%
\begin{tabular}{|c|r|r|r|r|r|r|r|r|r|r|r|r|r|r|r|r|r|r|r|r|r|r|r|r|r|r|r|r|}
\hline
\multirow{2}{*}{\textbf{Iter.}} & \multicolumn{4}{c|}{\textbf{Block1}} & \multicolumn{4}{c|}{\textbf{Block2}} & \multicolumn{4}{c|}{\textbf{Block3}} & \multicolumn{4}{c|}{\textbf{Block4}} & \multicolumn{4}{c|}{\textbf{Block5}} & \multicolumn{4}{c|}{\textbf{Block6}} & \multicolumn{4}{c|}{\textbf{Block7}} \\
\cline{2-29}
 & \textbf{E.} & \textbf{A.1} & \textbf{A.2} & \textbf{A.3} & \textbf{E.} & \textbf{A.1} & \textbf{A.2} & \textbf{A.3} & \textbf{E.} & \textbf{A.1} & \textbf{A.2} & \textbf{A.3} & \textbf{E.} & \textbf{A.1} & \textbf{A.2} & \textbf{A.3} & \textbf{E.} & \textbf{A.1} & \textbf{A.2} & \textbf{A.3} & \textbf{E.} & \textbf{A.1} & \textbf{A.2} & \textbf{A.3} & \textbf{E.} & \textbf{A.1} & \textbf{A.2} & \textbf{A.3} \\
\hline
\textbf{0} & \multicolumn{4}{c|}{244} & \multicolumn{4}{c|}{68} & \multicolumn{4}{c|}{89} & \multicolumn{4}{c|}{147} & \multicolumn{4}{c|}{68} & \multicolumn{4}{c|}{247} & \multicolumn{4}{c|}{765} \\
\hline
\textbf{1} & 145 & 208 & 145 & 192 & 29 & 30 & 29 & 43 & 14 & 12 & 14 & 62 & 125 & 96 & 125 & 140 & 38 & 42 & 38 & 58 & 78 & 153 & 78 & 159 & 394 & 675 & 394 & 759 \\
\hline
\textbf{2} & 233 & 201 & 198 & 165 & 18 & 26 & 18 & 30 & 5 & 4 & 6 & 51 & 112 & 87 & 77 & 89 & 39 & 35 & 27 & 47 & 49 & 137 & 73 & 66 & 374 & 613 & 368 & 759 \\
\hline
\textbf{3} & 217 & 198 & 124 & 122 & 14 & 37 & 17 & 29 & 2 & 2 & 0 & 21 & 112 & 63 & 43 & 61 & 4 & 25 & 20 & 30 & 38 & 128 & 47 & 48 & 353 & 588 & 355 & 758 \\
\hline
\textbf{4} & 129 & 199 & 124 & 115 & 16 & 26 & 17 & 18 & 0 & 0 & --- & 0 & 101 & 56 & 36 & 42 & 2 & 19 & 4 & 16 & 23 & 128 & 23 & 36 & 284 & 602 & 357 & 684 \\
\hline
\textbf{5} & 109 & 207 & 134 & 96 & 8 & 61 & 13 & 18 & --- & --- & --- & --- & 22 & 56 & 34 & 15 & 0 & 17 & 2 & 16 & 18 & 118 & 20 & 0 & 275 & 440 & 358 & 684 \\
\hline
$\Delta$ & -135 & -37 & -110 & \textcolor{teal}{\textbf{-148}} & \textcolor{teal}{\textbf{-60}} & -7 & -55 & -50 & \textcolor{teal}{\textbf{-89}} & \textcolor{teal}{\textbf{-89}} & \textcolor{teal}{\textbf{-89}} & \textcolor{teal}{\textbf{-89}} & -125 & -91 & -113 & \textcolor{teal}{\textbf{-132}} & \textcolor{teal}{\textbf{-68}} & -51 & -66 & -52 & -229 & -129 & -227 & \textcolor{teal}{\textbf{-247}} & \textcolor{teal}{\textbf{-490}} & -325 & -407 & -81 \\
\hline
-\% & -55.3 & -15.2 & -45.1 & \textcolor{teal}{\textbf{-60.7}} & \textcolor{teal}{\textbf{-88.2}} & -10.3 & -80.9 & -73.5 & \textcolor{teal}{\textbf{-100}} & \textcolor{teal}{\textbf{-100}} & \textcolor{teal}{\textbf{-100}} & \textcolor{teal}{\textbf{-100}} & -85.0 & -61.9 & -76.9 & \textcolor{teal}{\textbf{-89.8}} & \textcolor{teal}{\textbf{-100}} & -75.0 & -97.1 & -76.5 & -92.7 & -52.2 & -91.9 & \textcolor{teal}{\textbf{-100}} & \textcolor{teal}{\textbf{-64.1}} & -42.5 & -53.2 & -10.6 \\
\hline
\end{tabular}}
\par\vspace{2pt}
\parbox{2\columnwidth}{\raggedright
\textbf{E.} = EvoDRC, \textbf{A.1}--\textbf{A.3} = Ablation~1--Ablation~3. \textcolor{teal}{\textbf{Green}} numbers indicate the largest DRV count reduction for each block. the Dash marks iterations skipped by early stop after reaching zero DRVs.}
\vspace{-5pt}
\end{table*}

\begin{table}[t]
\centering
\vspace{-8pt}
\caption{EvoDRC LLM cost (\$) per agent per each block~\cite{claude-token-pricing}.}
\vspace{-12pt}
\fontsize{7pt}{8pt}\selectfont
\label{table:cost_full}
\begin{tabular}{|l|r|r|r|r|r|}
\hline
\textbf{Block} & \textbf{Repair} & \textbf{Skill Refiner} & \textbf{DB Summarizer} & \textbf{Skill Judge} & \textbf{Total} \\
\hline
Block1 & 144.57 & 9.33 & 10.85 & 3.01 & 167.76 \\
\hline
Block2 & 110.41 & 6.99 & 7.08 & 2.14 & 126.63 \\
\hline
Block3 & 50.45 & 7.11 & 6.26 & 1.97 & 65.79 \\
\hline
Block4 & 109.90 & 5.83 & 6.66 & 1.37 & 123.76 \\
\hline
Block5 & 65.71 & 9.04 & 6.69 & 2.69 & 84.14 \\
\hline
Block6 & 102.53 & 9.85 & 10.17 & 2.71 & 125.26 \\
\hline
Block7 & 411.76 & 13.70 & 25.29 & 8.14 & 458.89 \\
\hline
\textbf{Total} & 995.33 & 61.85 & 73.00 & 22.03 & 1152.23 \\
\hline
\textbf{\%} & 86.4\% & 5.4\% & 6.3\% & 1.9\% & 100.0\% \\
\hline
\end{tabular}
\par\vspace{2pt}
\parbox{\columnwidth}{\raggedright
Costs follow the Sonnet 4.6 API pricing per million tokens~\cite{claude-token-pricing}: \$3 base input, \$3.75 five-minute cache write, \$6 one-hour cache write, \$0.30 cache read, and \$15 output.}
\vspace{-9pt}
\end{table}

\providecommand{\SR}{\textbf{\textcolor{red}{S}}}
\providecommand{\DS}{\textbf{\textcolor{blue}{D}}}
\begin{table}[t]
\centering
\vspace{-6pt}
\caption{Majority skill file evolution winner across all metal layers for each block and iteration.}
\vspace{-12pt}
\fontsize{6pt}{8pt}\selectfont
\renewcommand{\arraystretch}{0.85}
\label{table:winner_all}
\setlength{\tabcolsep}{3pt}
\begin{tabular}{|c|c|c|c|c|c|c|c|c|c|c|c|c|c|c|c|c|c|c|c|c|c|}
\hline
\multirow{3}{*}{Iter} & \multicolumn{7}{c|}{\textbf{EvoDRC}} & \multicolumn{7}{c|}{\textbf{Ablation~1}} & \multicolumn{7}{c|}{\textbf{Ablation~3}} \\
\cline{2-22}
 & \multicolumn{7}{c|}{Block} & \multicolumn{7}{c|}{Block} & \multicolumn{7}{c|}{Block} \\
\cline{2-22}
 & 1 & 2 & 3 & 4 & 5 & 6 & 7 & 1 & 2 & 3 & 4 & 5 & 6 & 7 & 1 & 2 & 3 & 4 & 5 & 6 & 7 \\
\hline
1 & \DS & \DS & \SR & \DS & \DS & \DS & \DS & \DS & \DS & \DS & \DS & \DS & \DS & \DS & \DS & \DS & \DS & \DS & \DS & \DS & \DS \\
\hline
2 & \DS & T & \SR & \DS & T & T & \DS & \DS & \DS & \SR & \SR & \SR & \SR & \DS & \DS & \SR & \DS & \DS & \SR & \DS & -- \\
\hline
3 & \SR & \SR & \SR & -- & T & \SR & \SR & \DS & \SR & T & \SR & \SR & \SR & \DS & \DS & \DS & \DS & \DS & \SR & T & \DS \\
\hline
4 & \SR & \SR & \DS & \SR & \SR & \SR & \DS & \SR & \SR & \DS & T & \SR & \SR & \SR & T & \SR & \DS & \DS & \SR & \SR & \DS \\
\hline
\end{tabular}\\[2pt]
{\raggedright\SR{} = Skill Refiner, \DS{} = DB Summarizer; T = Tie, -- = no layer updated.\par}
\vspace{-9pt}
\end{table}

\subsection{Overall Repair Quality and Cost}
\label{subsec:results-overall}
\noindent
Table~\ref{table:results} shows the DRV count changes over iterations for EvoDRC and the three ablation studies. $\Delta$ indicates the change in DRV count relative to the initial count, $\%$ shows this change as a \% of the initial count. EvoDRC achieves an improvement of at least 55.3\% on all blocks, fully cleans Block3 and Block5, and reaches an 83.6\% improvement on average across all seven blocks, compared with 51.0\%, 77.9\%, and 73.0\% for Ablations~1--3. Table~\ref{table:cost_full} breaks down the EvoDRC dollar cost by agent role, using the Claude API pricing in effect at the time of publication~\cite{claude-token-pricing}. Compared with manually fixing hundreds of residual DRVs which can take a DRC engineer days, this is a small repair cost.

Ablation~1 confirms the value of skill transfer. Without the CLA-derived initial skill files, EvoDRC's improvement advantage reaches 77.9\% on Block2 and exceeds 20\% on every block except the DRC-clean Block3. On Block2, the ablation ends with more DRVs than it reaches at iteration~1. Because CLA is an external OpenCores design that is much smaller than the benchmark blocks and has a substantially different initial DRV distribution, the gap suggests that CLA-derived skills still help the first repair iterations on these targets. Ablation~2 confirms the value of skill evolution. With frozen skill files, its improvement is lower than EvoDRC on all six blocks that are not fully cleared, and its repair effectiveness drops in the late iterations. For example, Ablation~2 leaves two DRVs on Block5 while EvoDRC reaches a DRC-clean state, and it still leaves 358 DRVs on Block7 while EvoDRC further reduces the count to 275. Ablation~3 confirms that layout decomposition is essential for robust and scalable block-level repair. On Block7, EvoDRC removes 490 of 765 DRVs, whereas Ablation~3 removes only 81. This gap occurs because a single agent cannot cover enough violations within each iteration and leaves most violations unattempted. Repairing the entire design is competitive only on the smaller blocks. Therefore, decomposition is the mechanism that preserves EvoDRC's repair coverage across design scales, while the transferred initial skills and skill-evolution mechanisms determine the quality of repair.

\subsection{Skill Evolution Dynamics}
\label{subsec:results-distiller}

In practice, a DRC engineer often reuses repair strategies learned from earlier designs when a familiar violation pattern reappears, but must refine those strategies, or learn new ones, when a new layout exposes unseen rule interactions or geometric contexts. EvoDRC formulates this process as skill evolution. Experience from a previous design provides transferable initial skills, while verified repair outcomes on the target design continuously evolve them.

Each entry in Table~\ref{table:winner_all} reports the majority winner across all metal-layer skill updates for a specific experimental setting, block, and iteration. The winner is selected from the candidate skill-generation methods, namely the Skill Refiner and the DB Summarizer. Therefore, the table shows not only which method dominates overall, but also when the workflow switches from rewriting skills from newly collected records to refining already established skill files.

After iteration~1, the DB Summarizer wins almost every contest across EvoDRC, Ablation~1, and Ablation~3. At this stage, the newly collected repair records provide the first target-specific evidence for revising the skill files, and the DB Summarizer converts these records into updated repair knowledge, similar to how an engineer would draft a target-specific notebook from measured outcomes. From iteration~3 onward, the Skill Refiner, now operating on skill files that already contain target-specific knowledge, wins the majority of contests in all three configurations. 


\subsection{Skill Content Evolution}
\label{subsec:skill-content}

\begin{table}[t]
\centering
\vspace{-3pt}
\caption{Skill content evolution over iterations. Each row summarizes the content in the skill files for a repair iteration. }
\vspace{-12pt}
\fontsize{7pt}{10pt}\selectfont
\label{table:skill_content}
\resizebox{\columnwidth}{!}{%
\begin{tabular}{|c|r|r|r|r|r|r|r|r|r|r|r|r|}
\hline
\multirow{2}{*}{\textbf{Iter}} & \multicolumn{4}{c|}{\textbf{Block3}} & \multicolumn{4}{c|}{\textbf{Block5}} & \multicolumn{4}{c|}{\textbf{Block7}} \\
\cline{2-13}
    & \textbf{Cite} & \textbf{Sit.} & \textbf{Sug.} & \textbf{Warn} & \textbf{Cite} & \textbf{Sit.} & \textbf{Sug.} & \textbf{Warn} & \textbf{Cite} & \textbf{Sit.} & \textbf{Sug.} & \textbf{Warn} \\
\hline
1 & 0 & 25 & 24 & 46 & 0 & 25 & 24 & 46 & 0 & 25 & 24 & 46 \\
\hline
2 & 132 & 44 & 42 & 49 & 96 & 67 & 20 & 54 & 421 & 56 & 33 & 47 \\
\hline
3 & 207 & 38 & 42 & 56 & 230 & 70 & 28 & 67 & 498 & 77 & 31 & 57 \\
\hline
4 & 257 & 43 & 43 & 49 & 314 & 90 & 34 & 75 & 536 & 105 & 44 & 72 \\
\hline
5 & --- & --- & --- & --- & 405 & 95 & 33 & 75 & 525 & 108 & 62 & 81 \\
\hline
\end{tabular}}
\par\vspace{2pt}
\parbox{\columnwidth}{\raggedright
\textbf{Cite}: repair record citation count. \textbf{Sit.}: \# of unique situations defined by rule and layer geometry context. \# of \textbf{Sug.}: repair suggestions across situations. \textbf{Warn}: \# of warnings across situations.}

\vspace{-13pt}
\end{table}

\noindent
Each row in Table~\ref{table:skill_content} summarizes the skill files across all layers used in each iteration for Block3, Block5, and Block7\footnote{We show only results of three blocks for the interest of space. More skill files are available in~\cite{EvoDRC}}. A \textit{situation} is one DRC rule encountered in one unique layout geometry context, e.g., enclosure violation under a specific metal routing pattern, a \textit{suggestion} is one repair technique offered for a \textit{situation}, a \textit{warning} marks forbidden repair actions, and \textit{cite} is the reference repair record in the Knowledge DB.

The CLA-derived initial skill files describe 25 situations with 24 repair suggestions. There is no cited record at iteration~1 since the DRC repair has just started. Skill evolution then adapts the content to each iteration's repair trajectory. On Block5, the iteration-2 update prunes the suggestions from 24 to 20 after noticing the repair degradation, while situations and warnings keep growing to 95 and 75, so the skill file becomes more specific about when a fix applies and what it must not break. On Block7, situation coverage grows fastest, from 25 to 108, matching its largest and most diverse violation set. On Block3, which keeps clearing violations, suggestions accumulate to 43 while warnings stay low. This explicitly shows how evolution adapts the skill to each case's repair trajectory.

\section{Discussion}
\label{sec:discussion}

\begin{table}[t]
\centering
\vspace{-8pt}
\caption{Iteration traces of the EvoDRC runs on Block~3, Block~5, and Block~7 over representative rule types.}
\label{table:discussion_trace}
\vspace{-12pt}
\fontsize{6.5pt}{7pt}\selectfont
\begin{tabular}{|c|r|r|r|r|r|r|r|r|r|r|r|r|}
\hline
 & \multicolumn{3}{c|}{M1.S.2} & \multicolumn{3}{c|}{V2.M3.AUX.2} & \multicolumn{3}{c|}{M5.AUX.3} & \multicolumn{3}{c|}{V1.M1.EN.1} \\
\hline
Iter & B3 & B5 & B7 & B3 & B5 & B7 & B3 & B5 & B7 & B3 & B5 & B7 \\
\hline
0 & 5 & 6 & 25 & 27 & 21 & 225 & 0 & 0 & 0 & 6 & 5 & 26 \\
\hline
1 & 1 & 1 & 16 & 0 & 13 & 225 & 0 & 0 & 0 & 0 & 1 & 7 \\
\hline
2 & 0 & 0 & 12 & 0 & 6 & 225 & 0 & 13 & 0 & 0 & 0 & 4 \\
\hline
3 & 1 & 0 & 11 & 0 & 2 & 225 & 0 & 0 & 0 & 0 & 0 & 3 \\
\hline
4 & 0 & 0 & 11 & 0 & 0 & 153 & 0 & 0 & 0 & 0 & 0 & 5 \\
\hline
5 & --- & 0 & 11 & --- & 0 & 153 & --- & 0 & 0 & --- & 0 & 4 \\
\hline
\end{tabular}
\par\vspace{2pt}
\parbox{\columnwidth}{\raggedright
M1.S.2: minimum M1 tip-to-side spacing. V2.M3.AUX.2: V2 must match the M3 width perpendicular to the M3 direction. M5.AUX.3: M5 may not bend. V1.M1.EN.1: minimum enclosure of V1 by M1.}
\vspace{-16pt}
\end{table}

\noindent
Table~\ref{table:discussion_trace} traces the EvoDRC runs on Block3, Block5, and Block7, whose repair trajectories are shown via four representative rule types\footnote{We show only four rule types and three blocks for the interest of space. More information is available in~\cite{EvoDRC}}. M1.S.2 and V1.M1.EN.1 DRVs are cleaned in two iterations on Block3 and Block5, while Block7 keeps 11 of 25 M1.S.2 DRVs and 3 to 5 V1.M1.EN.1 DRVs after a first-iteration drop from 26 to 7. A repair introduces 13 M5.AUX.3 DRVs at iteration~2, degrading Block5 from 38 to 39 DRVs, and iteration~3 removes them with 90\% of the remainder. One iteration clears the 27 V2.M3.AUX.2 DRVs on Block3, and four iterations clear the 21 on Block5, while Block7 stays at 225 until a coordinated repair at iteration~4 clears 72.

Overall, the DRV count reaches zero on Block3 and Block5 within four iterations, while Block7 retains several residual DRVs, with any intermediate increases recovering within two iterations. This trend demonstrates the advantage of EvoDRC. Every repair operation, including those that temporarily increase the DRV count, is evaluated and recorded, allowing the agent to explore the consequences of its actions rather than a greedy search path. The knowledge evolved from these repair records captures the resulting collateral effects and enables more coordinated repairs in subsequent iterations.

\section{Conclusion}
\label{sec:conclusion}
\noindent
This paper presents EvoDRC, an agentic framework for post-route DRC repair. EvoDRC decomposes full-block layouts into bounded repair crops, repairs each crop with a four-role agent workflow, and records repair results. It further evolves layer-wise skill files through two paths, generated by the Skill Refiner agent and the DB Summarizer agent, respectively, and then selected by the Skill Judge agent. On DAC26-DRC-Benchmark, EvoDRC achieves an average DRV reduction of 83.6\%.

\clearpage
\bibliographystyle{misc/ieeetr2}
\bibliography{misc/bibfile}

\end{document}